\documentclass{article}

\usepackage{arxiv}

\usepackage[utf8]{inputenc} 
\usepackage[T1]{fontenc}    
\usepackage{hyperref}       
\usepackage{url}            
\usepackage{booktabs}       
\usepackage{amsfonts}       
\usepackage{nicefrac}       
\usepackage{microtype}      
\usepackage{natbib}         
\usepackage{lipsum}		
\usepackage{graphicx}
\usepackage{doi}
\usepackage{multirow}
\usepackage{colortbl}
\usepackage{dsfont}
\usepackage{amsmath,amssymb}
\usepackage{algorithm}
\usepackage{graphicx}
\usepackage{epstopdf}
\usepackage[caption=false]{subfig}
\usepackage{url} 
\usepackage{algpseudocode}
\usepackage{placeins}
\usepackage{tabularx}

\title{A Strong Balanced-Softmax Classifier-Retraining Baseline for Long-Tailed Recognition}

\date{July 10, 2026}	

\author{ \href{https://orcid.org/0000-0001-6662-0390}{\includegraphics[scale=0.06]{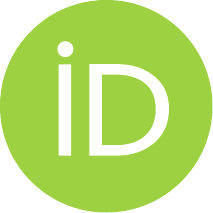}\hspace{1mm}Juan R.~Terven} \\
    CICATA Querétaro\\
	Instituto Politecnico Nacional\\
	Mexico \\
	\And
	\href{https://orcid.org/0000-0002-5657-7752}{\includegraphics[scale=0.06]{orcid.pdf}\hspace{1mm}Diana M.~Córdova-Esparza} \\
	Facultad de Informática\\
	Universidad Autónoma de Querétaro\\
	Mexico \\
	\And
    \href{https://orcid.org/0000-0001-7257-7595}{\includegraphics[scale=0.06]{orcid.pdf}\hspace{1mm}Julio Alejandro Romero-Gonzalez} \\
	Facultad de Informática\\
	Universidad Autónoma de Querétaro\\
	Mexico \\
    \And
    \href{https://orcid.org/0000-0002-1938-8610}{\includegraphics[scale=0.06]{orcid.pdf}\hspace{1mm}Edgar Arturo Chávez-Urbiola} \\
    CICATA Querétaro\\
	Instituto Politecnico Nacional\\
	Mexico \\
    \And
    \href{https://orcid.org/0000-0002-0979-2417}{\includegraphics[scale=0.06]{orcid.pdf}\hspace{1mm}Francisco Javier Willars-Rodríguez} \\
    CICATA Querétaro\\
	Instituto Politecnico Nacional\\
	Mexico \\
    \And
    \href{https://orcid.org/0000-0003-2663-2463}{\includegraphics[scale=0.06]{orcid.pdf}\hspace{1mm}Juan Bautista Hurtado Ramos} \\
    CICATA Querétaro\\
	Instituto Politecnico Nacional\\
	Mexico \\
    \And
    \href{https://orcid.org/0000-0003-0366-6249}{\includegraphics[scale=0.06]{orcid.pdf}\hspace{1mm}Alfonso Ramirez Pedraza} \\
    CICATA Querétaro\\
	Instituto Politecnico Nacional\\	
	Mexico \\
    \And
    \href{https://orcid.org/0000-0002-0315-1133}{\includegraphics[scale=0.06]{orcid.pdf}\hspace{1mm}Gendry Alfonso Francia} \\
    Escuela Nacional de Estudios Superiores\\
	Universidad Nacional Autónoma de México\\	
	Mexico \\
}

\hypersetup{
pdftitle={A Strong Balanced-Softmax Classifier-Retraining Baseline for Long-Tailed Recognition},
pdfsubject={cs, cs},
pdfauthor={Juan R.~Terven, Diana M.~Cordova-Esparza},
pdfkeywords={KEYWORD1, KEYWORD2},
}

\begin{document}
\maketitle

\begin{abstract}
Long-tailed recognition methods often modify losses, margins, or representations to reduce the dominance of frequent classes. We ask whether, after Balanced Softmax training, the remaining tail error can be reduced by retraining only the classifier. We evaluate BS-cRT, a two-stage procedure that trains a backbone and cosine classifier with Balanced Softmax, freezes the backbone, and updates only the classifier on balanced episodic batches. The second stage keeps the empirical-prior Balanced Softmax objective and uses raw cosine logits at inference. Across CIFAR-100-LT, CIFAR-10-LT, ImageNet-LT, and Places-LT, this classifier-only step consistently improves Few-shot accuracy over the matched Balanced Softmax checkpoint. At imbalance factor $100$,
Few-shot gains are $+5.15$ points on CIFAR-100-LT and $+5.83$ on CIFAR-10-LT;
on ImageNet-LT and Places-LT, gains are $+6.92$ and $+9.78$ points, respectively, with a Top-1/Few-shot trade-off on ImageNet-LT. We also analyze Counterfactual Boundary Risk Minimization (CBRM), a boundary-probe extension using prototype-based features near decision boundaries. CBRM identifies two failure modes: scaled-logit cosine margins destabilize training, and corrected hardest-negative probes remain head-class anchored. The results support BS-cRT as a practical classifier-side baseline and indicate that boundary supervision must account for class frequency.\end{abstract}


\section{Introduction}
\label{sec:intro}

Real-world visual recognition rarely enjoys the balanced label distributions assumed by standard benchmarks. Instead, categories typically follow a long-tailed frequency profile: a small number of head classes dominate the training set, while many tail classes appear only a handful of times. This imbalance is especially problematic because the rare classes are often the most important ones in practice, such as unusual medical findings, rare species, safety-critical failures, or infrequent industrial defects. Models trained naively on such data tend to allocate excessive decision space to head classes, yielding deceptively strong overall accuracy while failing on large parts of the label space~\citep{yang2022survey}.

A large body of work addresses long-tailed recognition by modifying the training signal induced by observed examples. Re-weighting methods amplify the loss of rare-class samples~\citep{cui2019classbalanced}; focal-style objectives reduce the contribution of easy examples~\citep{lin2017focal}; margin and logit-adjustment methods reshape the classifier geometry according to class frequency~\citep{cao2019ldam,ren2020balanced}; and decoupled training recipes separate representation learning from classifier learning, often retraining the classifier under a more balanced sampling distribution~\citep{kang2020decoupling}. Among these, Balanced Softmax has become a particularly strong and simple training-time baseline. At the same time, classifier retraining (cRT) has shown that much of the tail-class deficit can be addressed after a representation has already been learned.

This paper revisits the intersection of these two ideas. We ask a deliberately simple question: \emph{how far can classifier retraining go when the representation is trained with a strong Balanced Softmax anchor?} We instantiate this idea as BS-cRT, a minimal two-stage procedure. In stage 1, we train a backbone and cosine classifier end-to-end with Balanced Softmax. In stage 2, we freeze the backbone and retrain only the cosine classifier, again using Balanced Softmax, but now on class-balanced episodic batches containing $P$ classes and $K$ samples per class. The method has no prototypes, no synthetic samples, no auxiliary objectives, and no additional machinery beyond the usual classifier-retraining setup.

Despite its simplicity, BS-cRT is a strong baseline. Across CIFAR-100-LT and CIFAR-10-LT with imbalance factors $\{50,100,200\}$, BS-cRT consistently improves Few-shot accuracy over the same Balanced Softmax stage-1 checkpoint, while preserving or improving Top-1 accuracy. On CIFAR-100-LT at IF=100, BS-cRT improves Few-shot accuracy by $+5.15$ points over Balanced Softmax; on CIFAR-10-LT at IF=100, the gain is $+5.83$ points. The same pattern extends to larger-scale benchmarks: BS-cRT improves Few-shot accuracy by $+6.92$ points on ImageNet-LT and $+9.78$ points on Places-LT. Overall, the Few-shot gain is positive in every dataset--imbalance cell we evaluate, suggesting that a carefully matched classifier-retraining baseline remains a strong reference baseline.

We then ask a second question: \emph{can tail-class boundaries be improved further by injecting explicit synthetic supervision near decision boundaries?} This question motivated our initial method, Counterfactual Boundary Risk Minimization (CBRM). CBRM maintains multi-prototype class memories, constructs feature-space probes between a sample's positive prototype and a negative-class prototype, approximately bisects the current decision boundary, locally thickens the resulting probe, and penalizes cosine-margin violations using a class-balanced top-$k$ risk. The intuition is natural: tail classes have too few real samples to densely supervise their boundaries, so synthetic boundary-adjacent probes might provide additional corrective signal.

The experiments do not support this extension under hardest-negative probe selection, but they identify two useful failure modes. First, we identify a margin/scale mismatch that arises when CBRM is combined with cosine classifiers: the original implementation compared a margin parameter in the cosine-similarity range to logit differences scaled by the classifier temperature, producing gradients roughly $30\times$ larger than the Balanced Softmax anchor. We document this failure mode and fix it by computing CBRM violations directly in cosine-margin space. Second, even after this fix, hardest-negative CBRM probes do not improve over BS-cRT. On CIFAR-100-LT at IF=100, the corrected hardest-negative probe variant reduces Few-shot accuracy by approximately four points relative to BS-cRT. Our analysis suggests that this is not merely an optimization artifact: in long-tailed data, hardest negatives for tail samples are often head classes, so the generated probes are anchored near head-class regions and can pull tail classifier mass in the wrong direction.

For long-tailed recognition, the main implication is methodological. Balanced-Softmax classifier retraining provides a low-complexity way to improve tail accuracy without changing the representation, and it should be treated as a stage-2 control before tail gains are attributed to more complex mechanisms. The CBRM diagnostics also show that synthetic boundary supervision is not automatically beneficial: when probes are constructed from hardest negatives, they can be anchored by head-class competitors and can reinforce the bias that classifier retraining is meant to reduce. We therefore use BS-cRT as the main classifier-side baseline and report CBRM as a controlled negative result for a plausible boundary-supervision strategy.

The paper makes three contributions. First, it provides a matched-checkpoint evaluation of classifier retraining after Balanced Softmax, isolating the effect of a frozen-backbone classifier update from representation learning. Second, it shows that this stage-2 update consistently improves Few-shot accuracy across CIFAR-10/100-LT,
ImageNet-LT and Places-LT, with the expected trade-off between head-class contraction and tail-class recovery. Third, it analyzes a boundary-probe extension, CBRM, and identifies why hardest-negative probes can fail in long-tailed settings even after a cosine-margin correction. The code, configurations, and reporting scripts are released to support reproduction.

\section{Background and Related Work}

Long-tailed recognition lies at the intersection of statistical imbalance, representation learning, and classifier geometry. The central difficulty is not only that rare classes provide fewer examples, but that empirical risk minimization is driven by a skewed label distribution: head classes appear more often in mini-batches, contribute more gradient updates, and tend to induce larger and more stable decision regions~\citep{buda2018systematic,liu2019oltr,yang2022survey}. Tail classes, in contrast, must be recognized from limited evidence while competing against classifier directions learned from substantially richer support. This imbalance affects the learned features, the final classifier, posterior calibration, and the margins that separate classes in the embedding space \citep{cao2019ldam,kang2020decoupling,ren2020balanced,zhong2021improving}.

The literature has addressed this problem from several complementary angles. Some
methods rebalance the contribution of observed examples through resampling,
re-weighting, focal objectives, meta-learned weights, or effective-number corrections
\citep{chawla2002smote,lin2017focal,ren2018reweight,shu2019metaweight,cui2019classbalanced}.
Others modify the classifier objective directly through margins, logit adjustment,
prior-aware softmax normalizers, or label-distribution correction
\citep{cao2019ldam,menon2021logit,ren2020balanced,hong2021disentangling}.
A third family decouples representation learning from classifier learning, showing that
a useful feature extractor can still be paired with a biased classifier
\citep{kang2020decoupling,zhou2020bbn,tang2020causal,zhong2021improving}.
Recent work further explores contrastive objectives, prototype-based representations,
multi-expert models, and synthetic boundary supervision
\citep{cui2021parametric,zhu2022balanced,xuan2024dscl,snell2017prototypical,jian2025supervised,yang2026decision}.
These directions provide the context for our study, which revisits a deliberately simple
question: after training a representation with Balanced Softmax, how much tail accuracy
can be recovered by retraining only the classifier under balanced class exposure?

\subsection{Long-tailed recognition}

Visual recognition benchmarks have often been designed around balanced evaluation, but visual data collected in the wild rarely follows that pattern. Large image taxonomies, scene collections, and citizen-science species observations naturally allocate many samples to a small set of common categories and only a few samples to a large set of rare ones \citep{krizhevsky2009learning,deng2009imagenet,zhou2018places,vanhorn2018inat}. This imbalance is not only a data-management inconvenience. Under empirical risk minimization, head classes contribute more images, more mini-batch appearances, and more gradients. As a result, the learned posterior, the classifier weights, and often the confidence scores become correlated with the empirical training prior \citep{liu2019oltr,yang2022survey}. The common Many, Medium, and Few split used in long-tailed evaluation is therefore more than a reporting convention. It asks whether a method improves the neglected portion of the label space or merely preserves head-class dominance.

This paper follows that question to its smallest useful intervention. Rather than assuming that tail failure must be solved by a new backbone or a new representation objective, it asks how much rare-class accuracy remains latent in a representation already trained with a strong long-tailed loss. This makes the work close in spirit to the classical long-tailed recognition literature, but it places the spotlight on classifier geometry after Balanced Softmax training. The resulting position is deliberately strict: if a method improves Few-shot accuracy solely by retraining the final layer under more balanced exposure, then that gain should be credited to classifier retraining, not to a more elaborate mechanism.

\subsection{Rebalancing observed examples}

The most direct family of methods modifies the contribution of observed samples. Resampling changes which examples the optimizer sees, while re-weighting changes how strongly each observed loss affects the update. Systematic studies of class imbalance in convolutional networks showed that imbalance can strongly degrade recognition and that sampling choices are consequential \citep{buda2018systematic}. Focal Loss reduces the contribution of easy examples and became influential in dense detection, where the background and foreground imbalance is extreme \citep{lin2017focal}. Class-Balanced Loss replaces raw inverse-frequency weights with weights derived from the effective number of samples, which softens the correction when additional samples become redundant \citep{cui2019classbalanced}. Meta-learning approaches, including Learning to Reweight Examples and Meta-Weight-Net, go one step further by learning the weighting rule from validation feedback rather than fixing it by hand \citep{ren2018reweight,shu2019metaweight}.

These methods are important because they establish a central theme: long-tailed learning is not solved by treating every observed example equally. Yet they also expose a limitation that is central to the present paper. Re-weighting and re-sampling still act through the finite set of observed tail samples, and when used end-to-end, they can affect both the representation and the classifier at the same time. That coupling makes it harder to know whether a tail gain came from better features, from a better decision boundary, or from a changed confidence prior. BS-cRT avoids this ambiguity in stage 2. It freezes the representation and changes only the classifier, so the effect of balanced class exposure can be measured without re-learning the feature extractor.

\subsection{Prior-aware losses, margins, and logit correction}

A second line of work corrects the class-frequency bias directly in logit or margin space. LDAM gives rarer classes larger margins and combines this idea with deferred re-weighting, making the decision boundary itself class-frequency dependent \citep{cao2019ldam}. Logit adjustment provides a statistical view of long-tail learning by adding a label-prior correction to logits, either during training or after training \citep{menon2021logit}. Balanced Softmax reaches a closely related destination from the softmax normalizer: it incorporates the empirical class prior into the denominator so that optimization better matches a balanced evaluation protocol under label shift \citep{ren2020balanced}. LADE and PC Softmax further emphasize that the source label distribution can be disentangled from model predictions, and that long-tailed recognition can be read as a label-shift problem rather than only as a rare-sample problem \citep{hong2021disentangling}.

This group of methods is the immediate foundation for the present paper. Balanced Softmax is a strong anchor because it corrects the training prior without assigning explicit per-sample weights. However, a model trained with Balanced Softmax on naturally imbalanced mini-batches is still exposed to more head-class gradients than tail-class gradients. The loss changes the normalization geometry, but it does not make the stream of classifier updates class-balanced. BS-cRT tests precisely this residual bias: after prior-aware representation learning, can the classifier directions still be re-estimated more fairly by balanced episodic exposure?

Margin methods also motivate the diagnostic part of the paper. LDAM shows that margins should depend on class frequency, but normalized classifiers show that the unit of a margin matters. NormFace, CosFace, and ArcFace made hyperspherical feature and classifier geometry standard in face recognition by normalizing features and weights, then applying cosine or angular margins with an explicit scale parameter \citep{wang2017normface,wang2018cosface,deng2019arcface}. In such models, a margin expressed in cosine similarity is not numerically interchangeable with a margin computed after multiplying logits by the classifier scale. This is directly relevant to CBRM: a boundary-risk term must live in the same geometry as the cosine classifier. Otherwise, the intended margin becomes a scaled-logit penalty with a very different gradient magnitude.

\subsection{Decoupled representation and classifier learning}

The closest predecessor of BS-cRT is the decoupled training paradigm. Kang et al. showed that representation learning and classifier learning respond differently to imbalance, and that a representation learned with ordinary instance-balanced sampling can support strong long-tailed recognition once the classifier is re-estimated with class balancing \citep{kang2020decoupling}. Their classifier retraining strategy, cRT, along with weight-normalization variants such as tau-normalization and learnable weight scaling, shifted the field away from the assumption that every long-tail improvement must come from end-to-end rebalancing. BBN reached a similar conclusion through a two-branch architecture whose cumulative schedule separates broad representation learning from rebalanced classifier learning \citep{zhou2020bbn}. Causal Normalization interpreted part of the classifier bias through the momentum-induced confounding effect and removed the harmful head-class component at inference time \citep{tang2020causal}. MiSLAS added another important detail: two-stage long-tailed methods can improve accuracy while changing calibration, so classifier retraining should also be viewed as confidence redistribution \citep{zhong2021improving}.

Recent work continues to revisit this decoupled view rather than treating it as a solved baseline. Decoupled Optimization studies parameter groups that contribute differently to Many, Medium, and Few classes \citep{cong2024decoupled}. Label Over-Smooth rethinks classifier re-training under a unified feature representation and argues that simple label smoothing can balance the prediction space in the second stage \citep{sun2025rethinking}. These studies reinforce the premise that classifier learning is an active research axis, not just a post-processing footnote.

BS-cRT differs from classical cRT in two key ways. First, its first-stage representation is trained with Balanced Softmax instead of cross-entropy, making the stage-1 checkpoint a stronger, more relevant baseline. Second, its second-stage classifier is also trained with Balanced Softmax on balanced episodic batches, rather than switching to cross-entropy. This lets BS-cRT pose a sharper question than cRT: after a prior-aware loss shapes the representation, does balanced classifier exposure still recover tail accuracy? Empirically, yes—making BS-cRT a useful reference for evaluating more complex long-tailed methods.

\subsection{Representation learning, contrastive objectives, and prototypes}

A large body of complementary literature attempts to improve the representation itself. OLTR used memory-augmented features to transfer knowledge from head to tail classes in an open-world long-tailed setting \citep{liu2019oltr}. LEAP observed that head and tail classes occupy feature space differently, then augmented tail embeddings by transferring intra-class variation learned from head classes \citep{liu2020leap}. MetaSAug used meta-learning to learn semantic augmentation directions for minority classes, addressing the fact that direct covariance estimation is unreliable when a class has very few samples \citep{li2021metasaug}. DRO-LT framed the problem as representation bias and optimized a robustness-inspired loss to improve rare-class features \citep{samuel2021dro}. RIDE approached the same imbalance through multiple distribution-aware experts, reducing variance and bias with routing and diversity \citep{wang2021long}.

Contrastive learning brought another geometric perspective. Supervised Contrastive Learning pulls same-class samples together and pushes different-class samples apart in normalized embedding space \citep{khosla2020supcon}. In long-tailed data, however, the number of same-class positives and negative comparisons is itself imbalanced. Hybrid contrastive networks, PaCo, BCL, and Decoupled Contrastive Learning each address this issue by changing how positive pairs, class centers, or class-complement terms contribute to the contrastive objective \citep{wang2021hybrid,cui2021parametric,zhu2022balanced,xuan2024dscl}. Their analyses are highly relevant to this paper because they show that imbalance persists even in normalized representation spaces, where geometry is often easier to reason about.

Prototype-based methods connect representation geometry to classifier decisions. Prototypical Networks established a simple and powerful view of few-shot learning: classes can be represented by prototypes in an embedding space, and classification can be performed by distances to those prototypes \citep{snell2017prototypical}. CBRM borrows this geometric instinct but uses prototypes differently. It does not classify by prototype distance. Instead, it uses multi-prototype memories as anchors for synthetic boundary probes. The negative result in this paper therefore does not reject prototype structure itself. It shows that when prototypes are used to construct boundary supervision, the choice of negative prototype and the frequency profile of that negative class become part of the learning signal.

\subsection{Synthetic samples and boundary supervision}

The idea of adding synthetic evidence for minority classes predates deep long-tailed recognition. SMOTE creates synthetic minority samples by interpolation, while Borderline-SMOTE focuses on interpolation near minority examples that are likely to be misclassified \citep{chawla2002smote,han2005borderline}. In deep networks, mixup and Manifold Mixup generalized interpolation into input and hidden spaces, encouraging smoother behavior between training examples and flatter representations \citep{zhang2018mixup,verma2019manifold}. These methods support the intuition behind CBRM: if tail classes have too few samples to define reliable boundaries, then additional feature-space supervision near uncertain regions may help.

Recent long-tailed work has moved even closer to explicit boundary and decision-region supervision. Supervised Exploratory Learning synthesizes exploratory examples to improve class balance, decision regions, and decision boundaries without changing the backbone architecture \citep{jian2025supervised}. Decision Boundary-aware Generation uses generative augmentation to create informative near-boundary samples and reduce boundary ambiguity in long-tailed learning \citep{yang2026decision}. These works make boundary supervision a timely direction, but they also clarify why CBRM is a useful diagnostic. A synthetic boundary sample is not just another training point. It imposes a directional constraint on the classifier. If the negative anchor is usually a head class, as can happen with hardest-negative mining for tail samples, the synthetic probe can become head-anchored and can pull the tail classifier into the wrong region. In that regime, more boundary signal is not automatically better. It must be frequency-aware, geometrically consistent, and aligned to expand tail decision regions rather than restore head-class dominance.

Taken together, these lines of work motivate the design of our study. Prior-aware losses
and classifier retraining suggest that the final classifier should be examined separately
from the representation, while recent boundary-aware methods show that synthetic
near-boundary samples are a plausible way to target tail regions. BS-cRT tests the first
idea under a matched Balanced Softmax checkpoint. CBRM tests the second idea as a
controlled extension, and its negative result clarifies a failure mode of hardest-negative
probe construction in long-tailed data.

\clearpage

\section{Methods}
\label{sec:methods}

This section defines BS-cRT, the proposed classifier-retraining procedure, and then describes CBRM, the boundary-supervision objective evaluated as a diagnostic alternative. The notation is shared across both methods.

\subsection{Problem Setup}
\label{sec:method_setup}

Let $\mathcal{D}=\{(x_i,y_i)\}_{i=1}^{N}$ denote a long-tailed training set with labels $y_i \in \{1,\ldots,C\}$. The number of training examples in class $c$ is
\begin{equation}
    n_c = \left|\{i : y_i = c\}\right|,
\end{equation}
and the imbalance factor is $\mathrm{IF}=n_{\max}/n_{\min}$. A feature extractor $f_\theta:\mathcal{X}\rightarrow \mathbb{R}^{d}$ is followed by a cosine classifier with class weights $\{w_c\}_{c=1}^{C}$. Features and classifier weights are normalized before scoring:
\begin{equation}
    z = \frac{f_\theta(x)}{\lVert f_\theta(x)\rVert_2},
    \qquad
    \widehat{w}_c = \frac{w_c}{\lVert w_c\rVert_2},
    \qquad
    \ell_c(z)=s\,\widehat{w}_c^{\top}z,
\end{equation}
where $s$ is a fixed scale, set to $30$ in all experiments.

Balanced Softmax uses the empirical training prior
\begin{equation}
    \pi_c = \frac{n_c}{\sum_{c'=1}^{C} n_{c'}}
\end{equation}
and optimizes
\begin{equation}
    L_{\mathrm{BS}}(z,y)
    = -\log
    \frac{\exp\left(\ell_y(z)+\log \pi_y\right)}
    {\sum_{c=1}^{C}\exp\left(\ell_c(z)+\log \pi_c\right)}.
    \label{eq:balanced_softmax}
\end{equation}
At inference, we use the raw cosine logits
\begin{equation}
    \ell_c(z)=s\,\widehat{w}_c^{\top}z
\end{equation}
for prediction and confidence estimation. The log-prior term in Eq.~\ref{eq:balanced_softmax} is used only inside the training loss; it is not added to logits at test time. Thus, predictions are computed as $\arg\max_c \ell_c(z)$.

For reporting, classes are partitioned into Many-shot, Medium-shot, and Few-shot groups following standard long-tailed recognition protocols. On CIFAR-100-LT, for example, Many-shot classes satisfy $n_c>100$, Medium-shot classes satisfy $20<n_c\leq100$, and Few-shot classes satisfy $n_c\leq20$.

\subsection{BS-cRT: Balanced Softmax Classifier Retraining}
\label{sec:method_bscrt}

BS-cRT is a two-stage procedure. Stage 1 trains the backbone $f_\theta$ and cosine classifier end-to-end with Balanced Softmax on the original long-tailed training stream. We denote the resulting model by $f_{\theta}^{(1)}$ and $\{w_c^{(1)}\}_{c=1}^{C}$.

Stage 2 freezes $f_{\theta}^{(1)}$ and retrains only the cosine classifier. The stage-2 classifier is initialized from $\{w_c^{(1)}\}_{c=1}^{C}$, so the comparison isolates the effect of balanced classifier optimization from the same representation and checkpoint. Each stage-2 batch is sampled episodically. The sampler is configured with values $P$ and $K$, but the effective number of distinct classes per episode is $P_{\mathrm{eff}}=\min(P,C)$. We sample $P_{\mathrm{eff}}$ classes uniformly without replacement and then draw $K$ examples per selected class, yielding a batch of size $P_{\mathrm{eff}}K$. Within a selected class, examples are sampled without replacement when at least $K$ examples are available and with replacement otherwise. No additional sample weighting is applied. When $P_{\mathrm{eff}}<C$, the episodic batches provide a stochastic approximation to uniform class exposure across updates; when $P_{\mathrm{eff}}=C$, every class appears in each stage-2 batch. Optimization uses SGD with momentum $0.9$, weight decay $5\times10^{-4}$, and a cosine learning-rate schedule over $E_2$ epochs.

The stage-2 anchor loss remains Balanced Softmax with the empirical training prior. This point is important because the balanced sampler changes the outer sampling measure but does not redefine the prior inside the loss. In all BS-cRT experiments,
\begin{equation}
    \pi_c^{\mathrm{train}}
    = \frac{n_c}{\sum_{c'=1}^{C} n_{c'}}.
    \label{eq:stage2_empirical_prior}
\end{equation}
Thus, BS-cRT optimizes the same prior-aware per-sample surrogate as stage 1, but under balanced class exposure:
\begin{equation}
    L_{\mathrm{BS\text{-}cRT}}
    = \frac{1}{C}\sum_{c=1}^{C}
    \mathbb{E}_{x\sim \mathcal{D}_c}
    \left[
    -\log
    \frac{\exp\left(\ell_c(f_{\theta}^{(1)}(x))+\log \pi_c^{\mathrm{train}}\right)}
    {\sum_{j=1}^{C}\exp\left(\ell_j(f_{\theta}^{(1)}(x))+\log \pi_j^{\mathrm{train}}\right)}
    \right],
    \label{eq:bscrt_objective}
\end{equation}
where $\mathcal{D}_c$ denotes the empirical distribution over examples from class $c$. If one instead set $\pi_c=1/C$ in stage 2, the log-prior term would be constant across classes and would cancel from the softmax, reducing the anchor to ordinary cross-entropy under class-balanced sampling. That is not the objective evaluated in this work.

The rationale is that Balanced Softmax improves the stage-1 representation, but the classifier directions are still updated through an instance-balanced training stream in which head classes contribute more gradient steps. Stage 2 keeps the representation fixed and changes only the class exposure seen by the classifier. The empirical-prior Balanced Softmax anchor keeps the same prior-aware training objective as stage 1, while the balanced episodic sampler gives each selected class comparable influence on the retrained classifier directions. The BS-cRT objective contains no boundary-probe loss, no prototype-dependent gradient, and no auxiliary representation loss; in the implementation it is obtained by setting $\lambda_{\mathrm{cbr}} = 0$ and freezing the backbone.

\subsection{Practical Recipe and Cost}
\label{sec:method_practical_recipe}

Algorithm~\ref{alg:bscrt_recipe} summarizes the deployment procedure. The procedure requires only a Balanced Softmax checkpoint, the empirical class counts used by the loss, and the stage-2 hyperparameters. The procedure is fixed-schedule: the evaluation split is reserved for final reporting and is not used for checkpoint selection. BS-cRT uses no representation loss, prototype memory, or synthetic probe construction.

\begin{algorithm}[t]
\caption{Practical BS-cRT procedure. The loss uses the empirical training prior $\pi_c=n_c/\sum_{c'}n_{c'}$, while evaluation uses raw cosine logits.}
\label{alg:bscrt_recipe}
\small
\begin{algorithmic}[1]
\Statex \emph{Input:} Final scheduled Balanced Softmax checkpoint $(f_{\theta}^{(1)},\{w_c^{(1)}\})$, class counts $\{n_c\}$, $P$, $K$, learning rate, and number of stage-2 epochs.
\State Freeze the backbone $f_{\theta}^{(1)}$.
\State Initialize the cosine classifier from $\{w_c^{(1)}\}$.
\State Set $P_{\mathrm{eff}}=\min(P,C)$. At each update, sample $P_{\mathrm{eff}}$ classes uniformly without replacement and $K$ images per selected class, using replacement only when a selected class has fewer than $K$ examples.
\State Update only the classifier with $L_{\mathrm{BS}}$ using the empirical training prior.
\State Run the fixed stage-2 schedule and keep the final scheduled checkpoint, without selecting by evaluation-split Top-1.
\State Deploy the model with raw cosine logits $\ell_c(z)=s\widehat{w}_c^{\top}z$, without adding $\log \pi_c$ at inference.
\end{algorithmic}
\end{algorithm}

The computational footprint is small because only the classifier matrix is updated. With the normalized linear head used here, stage 2 trains $Cd$ parameters: $640$ parameters on CIFAR-10-LT, $6{,}400$ on CIFAR-100-LT, $2.05$M on ImageNet-LT, and $0.75$M on Places-LT. There is no additional inference-time module, memory, or latency relative to the stage-1 model. In our logs, one BS-cRT stage 2 run takes about $0.06$--$0.10$ hours on CIFAR-LT, $2.8$ hours on ImageNet-LT, and $2.2$ hours on Places-LT. On the two large-scale benchmarks, this corresponds to roughly $7$--$9\%$ of the corresponding stage-1 wall-clock time. BS-cRT therefore adds a short classifier-only training pass and leaves deployment unchanged.

Because stage 2 updates only the classifier, BS-cRT is suitable for deployment settings where the backbone is expensive to retrain or already validated. The inference graph is unchanged, so the method shifts the operating point toward rare classes without adding runtime modules, prototype memories, or latency. This makes the procedure especially relevant when rare categories carry operational value, for example, in defect detection, species recognition, or safety monitoring.

\subsection{CBRM as a Diagnostic Boundary-Probe Extension}
\label{sec:method_cbrm}

CBRM is evaluated as a diagnostic extension of BS-cRT rather than as the main method
of the paper. Its purpose is to test whether synthetic feature-space probes near current decision boundaries provide useful supervision beyond balanced classifier retraining.
All CBRM variants start from the same Balanced Softmax stage-1 checkpoint as
BS-cRT. The stage-2 objective is
\begin{equation}
    L =
    L_{\mathrm{BS}}
    + \lambda_{\mathrm{cbr}}L_{\mathrm{cbr}}
    + \lambda_{\mathrm{compact}}L_{\mathrm{compact}}
    + \lambda_{\mathrm{sep}}L_{\mathrm{sep}} .
    \label{eq:cbrm_full_objective}
\end{equation}
BS-cRT is recovered by setting $\lambda_{\mathrm{cbr}}=0$ and freezing the backbone.
The CBRM terms are included only to analyze whether boundary probes can improve on
this classifier-only baseline.

CBRM maintains $M$ unit-norm prototype buffers per class,
$\{\mu_{c,m}\}_{m=1}^{M}$. The prototypes are initialized from L2-normalized
features extracted by the stage-1 checkpoint and are updated during stage 2 by
exponential moving average, not by gradient descent. For a normalized feature $z$ and
class $c$, the nearest prototype slot is
\begin{equation}
    m^\star(z,c)=\arg\max_m z^\top \mu_{c,m}.
    \label{eq:cbrm_nearest_proto}
\end{equation}
If no samples are assigned to a prototype in a batch, that buffer is left unchanged.

For each stage-2 sample $(z_i,y_i)$, CBRM selects a negative class $j_i$. The default variant uses the current hardest negative,
\begin{equation}
    j_i = \arg\max_{k\neq y_i}\ell_k(z_i),
    \label{eq:cbrm_hard_negative}
\end{equation}
while the random-negative diagnostic samples $j_i$ uniformly from
$\{1,\ldots,C\}\setminus\{y_i\}$. The positive and negative prototype anchors are
\begin{equation}
    \mu_i^+ = \mu_{y_i,m^\star(z_i,y_i)},
    \qquad
    \mu_i^- = \mu_{j_i,m^\star(z_i,j_i)} .
    \label{eq:cbrm_anchors}
\end{equation}
CBRM then searches from the observed feature toward the negative prototype direction:
\begin{equation}
    d_i=\frac{\mu_i^- - \mu_i^+}{\|\mu_i^- - \mu_i^+\|_2},
    \qquad
    p_i(t)=\frac{z_i+t d_i}{\|z_i+t d_i\|_2},
    \qquad t\in[0,t_{\max}].
    \label{eq:cbrm_path}
\end{equation}
A fixed five-step bisection search seeks an approximate boundary point where
$\ell_{y_i}(p_i(t))-\ell_{j_i}(p_i(t))\approx0$. If the interval does not bracket a
crossing, the implementation keeps the endpoint reached by the search rather than
discarding the sample.

To avoid depending on a single boundary point, the probe is locally thickened by a
small normalized perturbation orthogonal to $d_i$. With $\xi\sim\mathcal{N}(0,I)$,
\begin{equation}
    \xi_\perp=\xi-(\xi^\top d_i)d_i,
    \qquad
    u_i=\frac{\xi_\perp}{\|\xi_\perp\|_2+\varepsilon},
    \qquad
    b_i=\frac{p_i(t_i^\star)+\sigma u_i}
    {\|p_i(t_i^\star)+\sigma u_i\|_2}.
    \label{eq:cbrm_thickened_probe}
\end{equation}

The corrected CBRM violation is computed in cosine-similarity space rather than on
scaled logits:
\begin{equation}
    v_i =
    \operatorname{softplus}\!\left(\gamma_i - m_i(b_i)\right),
    \qquad
    m_i(b)=\widehat{w}_{y_i}^{\top}b-\widehat{w}_{j_i}^{\top}b .
    \label{eq:cbrm_violation_compact}
\end{equation}
Here $m_i(b)\in[-2,2]$ is an unscaled cosine margin. The target margin is either fixed
or frequency-adaptive:
\begin{equation}
    \gamma_i=\gamma_0+\rho\log\frac{n_{\max}}{n_{y_i}}.
    \label{eq:cbrm_adaptive_margin_compact}
\end{equation}
In the reported runs, $\gamma_0=0.2$ and $\rho=0.05$. The largest configured value is
approximately $0.545$ on Places-LT, which remains inside the feasible cosine-margin
range. This cosine-space computation is important because applying the same target
margin to scaled logits would multiply the effective gradient with respect to the cosine
margin by the classifier scale $s=30$, which caused the loss spike observed in the
original CBRM diagnostic.

CBRM aggregates violations by class rather than by sample. Let $\mathcal{I}_c$ be the
set of probes in the batch whose true label is $c$. The classwise upper-tail risk and
boundary loss are
\begin{equation}
    r_c^{(\alpha)} =
    \frac{1}{\lceil \alpha |\mathcal{I}_c|\rceil}
    \sum_{i\in \operatorname{Top}_{\alpha}(\mathcal{I}_c)} v_i,
    \qquad
    L_{\mathrm{cbr}} =
    \frac{1}{|\mathcal{C}_{\mathrm{batch}}|}
    \sum_{c\in\mathcal{C}_{\mathrm{batch}}} r_c^{(\alpha)} .
    \label{eq:cbrm_class_balanced_risk_compact}
\end{equation}
We use $\alpha=0.25$; setting $\alpha=1$ recovers the classwise mean violation.

The compactness and separation terms in Eq.~\eqref{eq:cbrm_full_objective} are
prototype regularizers retained from the diagnostic implementation. They are not part
of BS-cRT. Since prototypes are EMA buffers, the separation term does not backpropagate
to the classifier or backbone in the reported implementation. The compactness term can
affect features only when part of the backbone is unfrozen; it is inert in the frozen-backbone
diagnostics.

Table~\ref{tab:stage2_variants} summarizes the stage-2 variants used in the diagnostic
experiments. The corrected CBRM variants compute violations in the cosine-margin
space of Eq.~\eqref{eq:cbrm_violation_compact}. The random-negative variant replaces
hardest-negative mining with uniform negative-class sampling to test whether probe
failures are tied to systematic head-class anchoring.

\begin{table}[t]
\centering
\small
\caption{Stage-2 variants evaluated in the diagnostic experiments. All variants start from the same Balanced Softmax stage-1 checkpoint for each seed.
}
\label{tab:stage2_variants}
\begin{tabular}{lllll}
\toprule
Variant & Trainable parameters & $\lambda_{\mathrm{cbr}}$ & Negative class & Boundary term \\
\midrule
BS-cRT & classifier only & 0 & none & none \\
CBRM original & classifier + final residual block & 1.0 & hardest & scaled-logit margin \\
CBRM fixed & classifier + final residual block & 1.0 & hardest & cosine margin \\
CBRM frozen + probes & classifier only & 1.0 & hardest & cosine margin \\
CBRM random negative & classifier only & 1.0 & random & cosine margin \\
\bottomrule
\end{tabular}
\end{table}

\subsection{Implementation Details}
\label{sec:method_implementation}

CIFAR-100-LT and CIFAR-10-LT use ResNet-32 at resolution $32\times32$. ImageNet-LT and Places-LT use ResNet-50 at resolution $224\times224$. All experiments use a normalized linear classifier with fixed scale $s=30$.

Stage 1 uses SGD with learning rate $0.1$, momentum $0.9$, weight decay $5\times10^{-4}$, batch size $128$, evaluation batch size $256$, and cosine learning-rate decay for $200$ epochs. CIFAR-LT uses no warmup, and AMP is disabled; ImageNet-LT and Places-LT use five warmup epochs, and AMP is enabled. To avoid test-set-based model selection, we use the checkpoint at the end of the fixed stage-1 schedule to initialize stage 2. The evaluation split is used only for final reporting.

Stage 2 initializes the classifier from the final scheduled stage-1 checkpoint,
freezes the backbone for BS-cRT, and trains for 40 epochs with SGD, learning rate
0.01, momentum 0.9, weight decay $5 \times 10^{-4}$, and cosine decay. The episodic
sampler is configured with $P=16$ and $K=8$ for all reported BS-cRT runs. In practice, the effective number of distinct classes per episode is $P_{\mathrm{eff}}=\min(P,C)$. Thus CIFAR-10-LT uses $P_{\mathrm{eff}}=10$ and batch
size $80$, while CIFAR-100-LT, ImageNet-LT, and Places-LT use $P_{\mathrm{eff}}=16$ and batch size $128$. CIFAR-LT uses 200 stage-2 batches per epoch, while ImageNet-LT and Places-LT use 500 batches per epoch. Evaluation uses batch size 256. The reported stage-2 model is the checkpoint at the end of this fixed schedule, not a best-by-evaluation checkpoint.

Training augmentations are the standard transforms used by the codebase. CIFAR-LT uses random crop with padding $4$, random horizontal flip, tensor conversion, and CIFAR normalization; evaluation uses tensor conversion and normalization only. ImageNet-LT and Places-LT use random resized crop to $224$, random horizontal flip, color jitter with brightness, contrast, and saturation strength $0.4$, and ImageNet normalization; evaluation resizes to $256$, center-crops to $224$, and applies the same normalization.

Unless otherwise specified, CBRM uses $\lambda_{\mathrm{cbr}}=1.0$, $\lambda_{\mathrm{compact}}=0.05$, and $\lambda_{\mathrm{sep}}=0.01$. The no-probe BS-cRT variants set $\lambda_{\mathrm{cbr}}=0$; in frozen-backbone runs, the compactness and separation terms do not update trainable parameters, as described above. CBRM also uses five bisection steps, $t_{\max}=0.8$, $M=2$ prototypes per class, EMA coefficient $\alpha_{\mathrm{ema}}=0.99$, tangent noise scale $\sigma_0=0.02$, top-tail fraction $\alpha=0.25$, base margin $\gamma_0=0.2$, frequency-adaptive margin coefficient $\rho=0.05$, and a three-epoch warmup before enabling the boundary-risk term. All main results are reported over seeds $\{1,2,3\}$ with seeded data loaders and checkpointed configurations.

All experiments were run on a local workstation equipped with two NVIDIA GeForce
RTX 4090 GPUs, an Intel Core i9-13900K CPU, and 128 GB of system memory. Each
training run used a single GPU unless otherwise stated. The wall-clock times reported
in Section~\ref{sec:method_practical_recipe} were measured on this workstation.

\section{Results}
\label{sec:experiments}

The experiments first measure the effect of BS-cRT under matched stage-1 checkpoints, then examine whether the effect persists across imbalance factors and larger benchmarks. The final part of the section analyzes CBRM to determine whether synthetic boundary probes add useful supervision beyond classifier retraining. Unless
otherwise stated, all values are percentages averaged over three seeds. Balanced Softmax is the primary reference because BS-cRT starts from the same stage-1 checkpoint and changes only the stage-2 classifier optimization.

\subsection{CIFAR-LT at IF 100}
\label{sec:results_cifar_if100}

Table~\ref{tab:cifar_if100_results} reports the central CIFAR comparison at an imbalance factor of $100$. On CIFAR-100-LT, BS-cRT approximately preserves Top-1 accuracy, moving from $40.73$ to $41.07$, while increasing Few-shot accuracy from $19.91$ to $25.06$, a gain of $+5.15$ points over Balanced Softmax. On CIFAR-10-LT, BS-cRT improves Top-1 from $75.65$ to $77.82$ and Few-shot accuracy from $64.34$ to $70.17$, a gain of $+5.83$ points. These gains are not a consequence of a stronger representation, since the backbone is frozen and shared with the Balanced Softmax stage-1 model.

\begin{table*}[t]
\centering
\caption{Main CIFAR-LT results at imbalance factor $100$. Top-1 is reported as mean $\pm$ standard deviation over three seeds. Many, Medium, and Few are class-frequency bucket accuracies. All values are percentages.}
\label{tab:cifar_if100_results}
\setlength{\tabcolsep}{4pt}
\small
\begin{tabular}{lrrrrrrrr}
\toprule
& \multicolumn{4}{c}{CIFAR-100-LT} & \multicolumn{4}{c}{CIFAR-10-LT} \\
\cmidrule(lr){2-5} \cmidrule(lr){6-9}
Method & Top-1 & Many & Medium & Few & Top-1 & Many & Medium & Few \\
\midrule
CE & $37.33\pm0.67$ & 65.89 & 36.16 & 9.86 & $71.25\pm0.98$ & 89.58 & 70.28 & 53.40 \\
Focal & $28.28\pm1.76$ & 52.06 & 25.26 & 7.34 & $67.62\pm1.87$ & 85.56 & 62.47 & 52.27 \\
LDAM-DRW & $39.82\pm0.14$ & 65.47 & 39.56 & 14.40 & -- & -- & -- & -- \\
Balanced Softmax & $40.73\pm0.96$ & 61.85 & 40.42 & 19.91 & $75.65\pm0.55$ & 88.77 & 72.02 & 64.34 \\
cRT & $39.57\pm0.70$ & 64.66 & 39.27 & 14.75 & $74.43\pm1.48$ & 87.71 & 72.37 & 62.19 \\
BS-cRT (ours) & $41.07\pm1.01$ & 57.16 & 40.99 & 25.06 & $77.82\pm0.18$ & 87.33 & 74.10 & 70.17 \\
\bottomrule
\end{tabular}
\end{table*}

\begin{figure*}[t]
\centering
\includegraphics[width=0.92\textwidth]{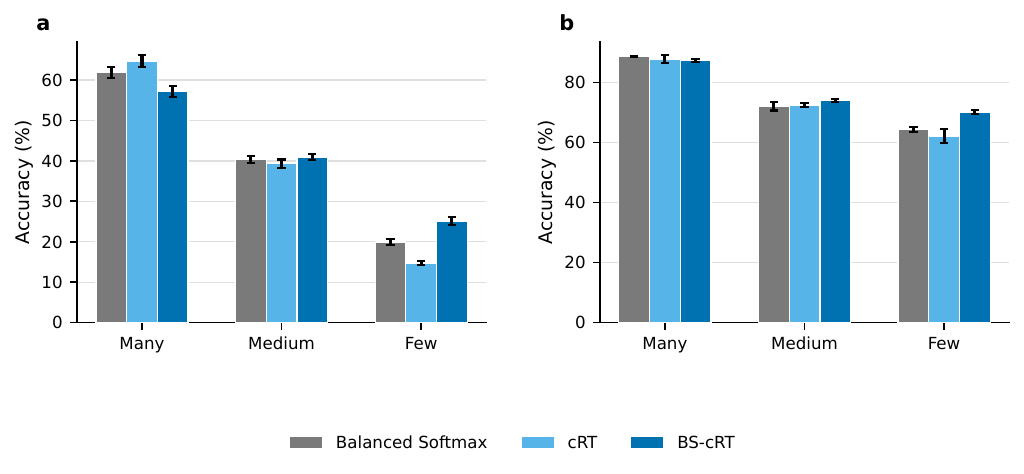}
\caption{Frequency-bucket breakdown on CIFAR-LT at imbalance factor $100$. Panel (a) shows CIFAR-100-LT, and panel (b) shows CIFAR-10-LT. The figure compares Balanced Softmax, classical cRT, and BS-cRT on Many, Medium, and Few classes. BS-cRT trades a moderate reduction in Many-shot accuracy for a larger gain on Few-shot classes, which is the desired behavior in balanced long-tailed evaluation. Error bars show standard deviation across three seeds.}
\label{fig:bucket_if100}
\end{figure*}

Figure~\ref{fig:bucket_if100} shows the same result from the frequency-bucket perspective. BS-cRT gives up some Many-shot accuracy, as expected from class-balanced classifier retraining, but it recovers substantially more accuracy on the underrepresented classes. On CIFAR-100-LT, Few-shot accuracy rises by more than five points while Medium accuracy is preserved. On CIFAR-10-LT, the shift is even cleaner: Medium and Few improve together, and the Many-shot contraction remains small enough that Top-1 also improves. This bucket-level behavior is the core empirical signature of BS-cRT: it reallocates classifier capacity toward the tail while retaining enough head-class performance to avoid a global accuracy collapse.

\subsection{Consistency Across Benchmark Cells}
\label{sec:results_consistency}

\begin{figure*}[t]
\centering
\includegraphics[width=\textwidth]{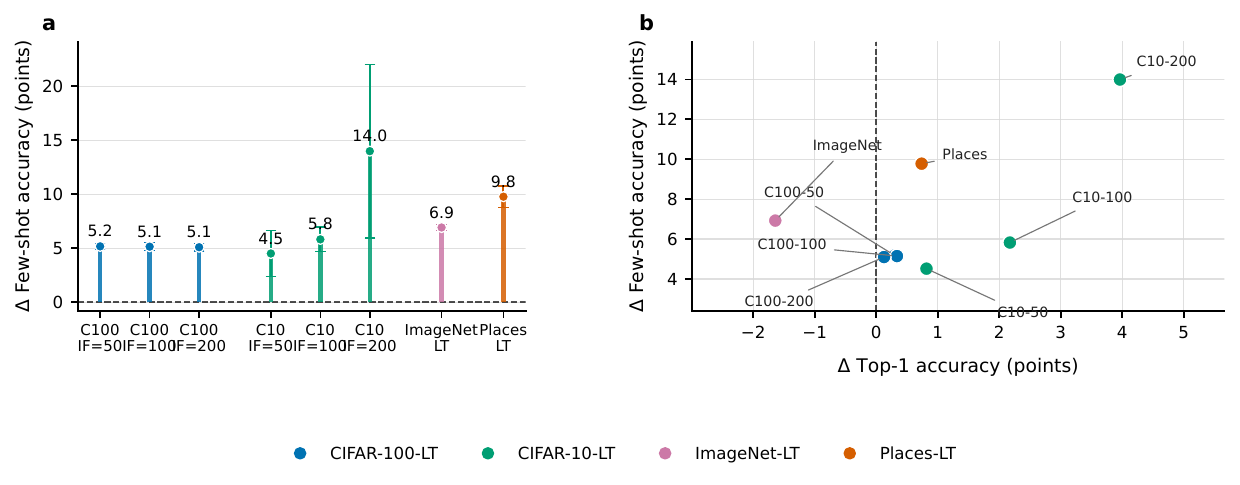}
\caption{Summary of BS-cRT improvements over Balanced Softmax. Panel (a) shows the paired seedwise Few-shot gain, $\Delta_{\mathrm{Few}} = \mathrm{Few}_{\mathrm{BS\text{-}cRT}} - \mathrm{Few}_{\mathrm{BS}}$, for each benchmark cell. Error bars denote the standard deviation of paired seed differences. Panel (b) plots the corresponding Top-1 change against the Few-shot change. Labels use compact dataset and imbalance-factor notation, for example, C100-50 denotes CIFAR-100-LT IF $50$. All evaluated cells lie above zero on the Few-shot axis, while Top-1 is preserved or improved in all but ImageNet-LT.}
\label{fig:bscrt_summary}
\end{figure*}

The same pattern appears across imbalance factors and datasets. Figure~\ref{fig:bscrt_summary} summarizes paired seedwise differences between BS-cRT and Balanced Softmax across all evaluated benchmark cells, and Table~\ref{tab:paired_fewshot_ci} reports the corresponding seedwise Few-shot deltas with paired $95\%$ confidence intervals. BS-cRT improves Few-shot accuracy for every seed in every evaluated benchmark cell: the mean gains are $+5.18$, $+5.15$, and $+5.10$ points on CIFAR-100-LT for imbalance factors $50$, $100$, and $200$; $+4.52$, $+5.83$, and $+13.99$ points on CIFAR-10-LT; $+6.92$ points on ImageNet-LT; and $+9.78$ points on Places-LT. The Top-1 columns in Table~\ref{tab:paired_fewshot_ci} clarify the operating point: CIFAR-100-LT is best read as Top-1 preservation with tail recovery, CIFAR-10-LT and Places-LT show Top-1 gains, and ImageNet-LT is a clear Top-1/Few-shot trade-off.

\begin{table*}[t]
\centering
\caption{Paired BS-cRT gains over the matched Balanced Softmax checkpoint. Seed columns report $\Delta_{\mathrm{Few}}$ in percentage points for seeds $1$, $2$, and $3$. Confidence intervals are two-sided paired $t$ intervals over the three seedwise differences. Because each interval is computed from three paired seeds, the confidence intervals are descriptive and should be interpreted together with the seedwise differences. The Top-1 column is included to make the operating point explicit rather than to claim uniform aggregate improvement.}
\label{tab:paired_fewshot_ci}
\setlength{\tabcolsep}{4pt}
\small
\resizebox{\textwidth}{!}{%
\begin{tabular}{lrrrrr}
\toprule
Benchmark cell & Seed 1 & Seed 2 & Seed 3 & Mean $\Delta_{\mathrm{Few}}$ [95\% CI] & Mean $\Delta$Top-1 [95\% CI] \\
\midrule
CIFAR-100-LT IF $50$  & $+4.97$ & $+5.09$ & $+5.47$ & $+5.18$ [$+4.53$, $+5.83$] & $+0.30$ [$+0.13$, $+0.47$] \\
CIFAR-100-LT IF $100$ & $+5.38$ & $+5.35$ & $+4.71$ & $+5.15$ [$+4.20$, $+6.10$] & $+0.34$ [$+0.13$, $+0.54$] \\
CIFAR-100-LT IF $200$ & $+4.82$ & $+4.97$ & $+5.50$ & $+5.10$ [$+4.21$, $+5.98$] & $+0.13$ [$-0.79$, $+1.04$] \\
CIFAR-10-LT IF $50$   & $+2.25$ & $+6.50$ & $+4.80$ & $+4.52$ [$-0.80$, $+9.83$] & $+0.82$ [$+0.15$, $+1.48$] \\
CIFAR-10-LT IF $100$  & $+5.63$ & $+7.05$ & $+4.80$ & $+5.83$ [$+3.00$, $+8.65$] & $+2.17$ [$+1.06$, $+3.29$] \\
CIFAR-10-LT IF $200$  & $+15.50$ & $+21.18$ & $+5.30$ & $+13.99$ [$-5.99$, $+33.97$] & $+3.96$ [$-0.21$, $+8.14$] \\
ImageNet-LT            & $+6.75$ & $+7.28$ & $+6.75$ & $+6.92$ [$+6.17$, $+7.68$] & $-1.64$ [$-2.19$, $-1.10$] \\
Places-LT              & $+10.59$ & $+10.07$ & $+8.67$ & $+9.78$ [$+7.31$, $+12.24$] & $+0.74$ [$-0.29$, $+1.76$] \\
\bottomrule
\end{tabular}%
}
\end{table*}

We additionally report non-parametric robustness checks over the eight evaluated benchmark cells in Table~\ref{tab:nonparametric_checks}. These checks use the cell-level mean deltas from Table~\ref{tab:paired_fewshot_ci}, rather than treating all
seed-level runs as independent observations. Few-shot gains are positive in all eight cells, yielding $p=0.0078$ under exact sign, Wilcoxon signed-rank, and sign-flip tests. Top-1 changes are positive in seven of eight cells but do not reject the null under the same checks, consistent with the ImageNet-LT Top-1/Few-shot trade-off.

\begin{table}[t]
\centering
\small
\caption{
Non-parametric robustness checks for BS-cRT gains over the matched Balanced
Softmax checkpoint. Tests are computed over the eight evaluated benchmark cells
using the cell-level mean deltas from Table~\ref{tab:paired_fewshot_ci}. Mean and median deltas are reported in percentage points. All tests are two-sided and are intended as robustness checks rather than replacements for the seedwise paired deltas.
}
\label{tab:nonparametric_checks}
\begin{tabular}{lcccccc}
\toprule
Metric &
Mean $\Delta$ &
Median $\Delta$ &
Positive cells &
Sign test $p$ &
Wilcoxon $p$ &
Sign-flip $p$ \\
\midrule
Few-shot & $+7.06$ & $+5.51$ & $8/8$ & $0.0078$ & $0.0078$ & $0.0078$ \\
Top-1    & $+0.85$ & $+0.54$ & $7/8$ & $0.0703$ & $0.1094$ & $0.2031$ \\
\bottomrule
\end{tabular}
\end{table}

\begin{figure*}[t]
\centering
\includegraphics[width=0.92\textwidth]{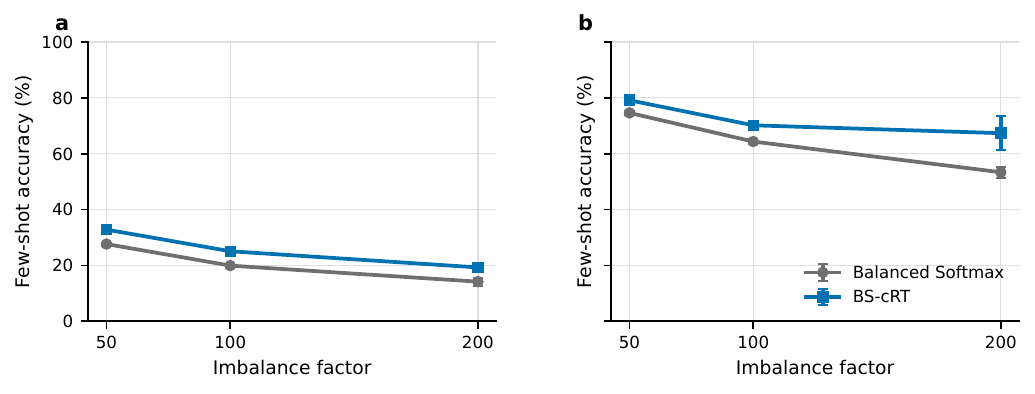}
\caption{Few-shot accuracy as a function of imbalance factor. Panel (a) shows CIFAR-100-LT and panel (b) shows CIFAR-10-LT, using a shared vertical scale from $0$ to $100$ for direct comparison. BS-cRT is compared against the Balanced Softmax checkpoint from which it is initialized. Error bars show standard deviation across three seeds. The gap is positive at every imbalance factor and grows sharply for CIFAR-10-LT at IF $200$.}
\label{fig:cifar_if_sweep}
\end{figure*}

Figure~\ref{fig:cifar_if_sweep} isolates the role of imbalance severity on CIFAR. On CIFAR-100-LT, the Few-shot gain remains stable as the imbalance factor increases: $+5.18$, $+5.15$, and $+5.10$ points for IF $50$, $100$, and $200$. On CIFAR-10-LT, the gain grows with imbalance and reaches $+13.99$ points at IF $200$. This supports the mechanism behind BS-cRT. When stage-1 training is increasingly dominated by head-class gradients, re-estimating classifier directions under balanced class exposure becomes increasingly valuable.

\subsection{Large-Scale Benchmarks}
\label{sec:results_large_scale}

Table~\ref{tab:large_scale_results} reports ImageNet-LT and Places-LT. On ImageNet-LT, BS-cRT should be interpreted as a Top-1/Few-shot trade-off: Top-1 decreases from $51.01$ to $49.37$, while Few-shot accuracy rises substantially from $38.21$ to $45.13$. On Places-LT, where the natural imbalance is more extreme, the same classifier-retraining procedure improves both Top-1 and Few-shot accuracy. Top-1 increases from $27.64$ to $28.38$, and Few-shot accuracy rises from $18.61$ to $28.39$.

\begin{table}[!b]
\centering
\caption{Large-scale benchmark results. Top-1 is reported as mean $\pm$ standard deviation over three seeds. Many, Medium, and Few are class-frequency bucket accuracies. All values are percentages.}
\label{tab:large_scale_results}
\setlength{\tabcolsep}{4pt}
\small
\begin{tabular}{lrrrrrrrr}
\toprule
& \multicolumn{4}{c}{ImageNet-LT} & \multicolumn{4}{c}{Places-LT} \\
\cmidrule(lr){2-5} \cmidrule(lr){6-9}
Method & Top-1 & Many & Medium & Few & Top-1 & Many & Medium & Few \\
\midrule
CE & $46.00\pm1.16$ & 68.82 & 47.00 & 22.42 & $21.41\pm0.18$ & 39.23 & 18.51 & 6.58 \\
Balanced Softmax & $51.01\pm0.61$ & 63.56 & 51.39 & 38.21 & $27.64\pm0.30$ & 36.84 & 27.54 & 18.61 \\
cRT & $49.36\pm0.94$ & 65.16 & 50.53 & 32.54 & $25.94\pm0.55$ & 38.96 & 25.25 & 13.70 \\
BS-cRT (ours) & $49.37\pm0.80$ & 54.49 & 48.51 & 45.13 & $28.38\pm0.14$ & 26.24 & 30.54 & 28.39 \\
\bottomrule
\end{tabular}
\end{table}

\begin{figure*}[!b]
\centering
\includegraphics[width=0.92\textwidth]{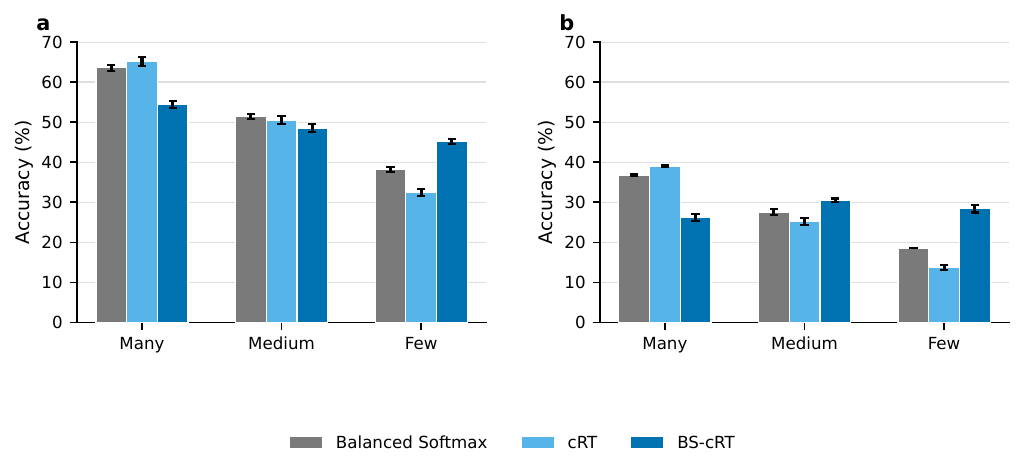}
\caption{Frequency-bucket breakdown on ImageNet-LT and Places-LT. Panel (a) shows ImageNet-LT and panel (b) shows Places-LT, using a shared vertical scale from $0$ to $70$ for direct comparison. The same classifier-retraining recipe produces a strong Few-shot gain on both datasets. ImageNet-LT exposes a Top-1 trade-off because the Many-shot contraction is large enough to outweigh the Few-shot gain in the aggregate. In contrast, on Places-LT the Medium- and Few-shot gains compensate for the Many-shot reduction. Error bars show standard deviation across three seeds.}
\label{fig:large_scale_buckets}
\end{figure*}

Figure~\ref{fig:large_scale_buckets} makes the large-scale trade-off explicit. On ImageNet-LT, the Few-shot gain comes with a visible Many-shot contraction, so BS-cRT should be interpreted as a tail-favoring operating point rather than a universal Top-1 improvement. On Places-LT, the same rebalancing improves Medium and Few classes enough to increase overall Top-1 despite a reduction in Many-shot classes. In both cases, the method behaves consistently with its design: it reduces the classifier's reliance on head-class decision regions.

Table~\ref{tab:published_context} places our results next to representative published results under common long-tailed recognition protocols. The cited methods differ in backbone, training length, augmentation policy, auxiliary objectives, generative or contrastive branches, and in some cases model multiplicity. The controlled evidence for BS-cRT is therefore the matched-checkpoint comparison in Tables~\ref{tab:cifar_if100_results} and~\ref{tab:large_scale_results}. The role of Table~\ref{tab:published_context} is to show where this low-complexity classifier-side correction sits relative to stronger published systems.

\begin{table*}[t]
\centering
\scriptsize
\setlength{\tabcolsep}{4pt}
\caption{
Published results context under common long-tailed recognition protocols.
All values are top-1 accuracy percentages. ``All'' denotes overall accuracy,
and ``Few'' denotes the Few-shot/Tail split reported by the corresponding paper.
The rows are not directly comparable because published methods use different
backbones, training schedules, augmentation policies, auxiliary objectives, or
model multiplicity. The table is intended as literature context, not as a controlled comparison.
}
\label{tab:published_context}
\resizebox{\textwidth}{!}{%
\begin{tabular}{llcccccc}
\toprule
Method &
Main mechanism &
\multicolumn{2}{c}{CIFAR-100-LT IF=100} &
\multicolumn{2}{c}{ImageNet-LT} &
\multicolumn{2}{c}{Places-LT} \\
\cmidrule(lr){3-4}\cmidrule(lr){5-6}\cmidrule(lr){7-8}
& & All & Few/Tail & All & Few & All & Few \\
\midrule
cRT~\citep{kang2020decoupling} &
Classifier retraining &
43.3 & 18.1 & 49.6 & 27.3 & 36.7 & 24.9 \\

BALMS / Balanced Softmax~\citep{ren2020balanced} &
Prior-aware softmax &
50.8 & -- & 41.8 & 25.3 & 38.7 & 31.6 \\

BSCE / Balanced Softmax~\citep{ren2020balanced,cong2024decoupled} &
Prior-aware loss &
50.8 & 33.4 & 52.3 & 33.4 & 39.4 & 32.7 \\

MiSLAS~\citep{zhong2021improving} &
Two-stage learning and calibration &
46.8 & 26.6 & 51.4 & 32.8 & 37.6 & 27.5 \\

LADE~\citep{hong2021disentangling} &
Label-distribution disentangling &
45.6 & 29.8 & 52.3 & 34.3 & 39.2 & 32.3 \\

PaCo~\citep{cui2021parametric} &
Parametric contrastive learning &
52.0 & -- & 57.0 & 39.1 & 41.2 & 35.3 \\

RIDE~\citep{wang2021long} &
Multi-expert routing &
48.0 & 23.7 & 55.7 & 36.0 & 40.3 & 33.0 \\

BCL~\citep{zhu2022balanced} &
Balanced contrastive learning &
53.9 & 35.5 & 57.1 & 36.6 & -- & -- \\

GCL~\citep{li2022gcl} &
Logit/feature rebalancing &
48.7 & -- & 54.9 & -- & -- & -- \\

Decoupled Optimisation~\citep{cong2024decoupled} &
Grouped parameter optimization &
53.8 & 44.6 & 60.4 & 52.2 & 42.8 & 40.1 \\

Label Over-Smooth~\citep{sun2025rethinking} &
Classifier retraining regularisation &
54.9 & -- & 54.4 & 42.3 & -- & -- \\

GLMC+SEL~\citep{du2023global,jian2025supervised} &
Exploratory synthetic examples &
56.48 & -- & 57.24 & 38.28 & -- & -- \\

CBDM+DBG~\citep{yang2026decision} &
Boundary-aware generation &
51.21 & 30.10 & -- & -- & -- & -- \\

\midrule
BS-cRT (ours) &
Classifier-only retraining after Balanced Softmax &
$41.07 \pm 1.01$ & 25.06 &
$49.37 \pm 0.80$ & 45.13 &
$28.38 \pm 0.14$ & 28.39 \\
\bottomrule
\end{tabular}
}
\vspace{1mm}
\begin{minipage}{0.98\textwidth}
\footnotesize
\emph{Notes.}
Dashes indicate that the corresponding split or dataset was not reported. BALMS / Balanced Softmax denotes the original BALMS report, whereas BSCE / Balanced Softmax denotes the Balanced Softmax baseline reported in the Decoupled Optimisation comparison.
\end{minipage}
\end{table*}

This comparison also defines the intended scope of BS-cRT. On CIFAR-100-LT,
modern representation-learning, multi-stage, and augmentation-heavy methods achieve
higher overall accuracy. On ImageNet-LT, BS-cRT gives a strong Few-shot operating
point despite updating only the classifier, but this comes with the Top-1 reduction
reported above. The main point is not that BS-cRT replaces heavily engineered systems,
but that those systems should control for a matched classifier-retraining stage before
attributing tail gains to additional mechanisms.

\subsection{CBRM Diagnostics}
\label{sec:results_cbrm}

\begin{figure*}[t]
\centering
\includegraphics[width=\textwidth]{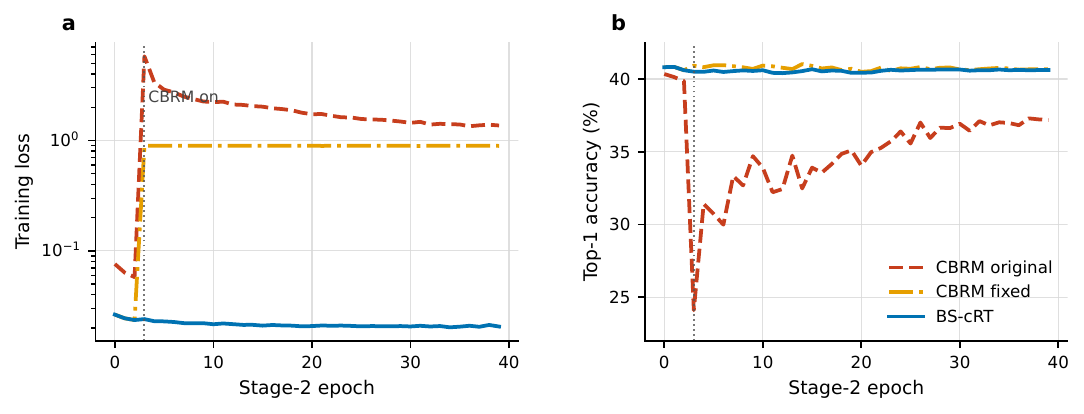}
\caption{Training dynamics of CBRM and BS-cRT on CIFAR-100-LT IF $100$, seed $1$. Panel (a) shows stage-2 training loss on a logarithmic scale. Panel (b) shows Top-1 accuracy during the same run. The original CBRM formulation exhibits a sharp loss spike and accuracy collapse when the boundary-risk term activates. The cosine-margin fix stabilizes optimization, but BS-cRT remains the more reliable stage-2 baseline.}
\label{fig:cbrm_diagnostic}
\end{figure*}

We next evaluate CBRM as a boundary-supervision extension of BS-cRT. Under the tested hardest-negative construction, the extension does not improve the classifier-only baseline. The original CBRM implementation suffers from a margin/scale mismatch when used with a cosine classifier: the target margin is configured in cosine-similarity units, while the violation is computed on logits multiplied by the classifier scale. Figure~\ref{fig:cbrm_diagnostic} shows the consequence on CIFAR-100-LT IF $100$, seed $1$. When CBRM activates after warmup, training loss jumps by nearly two orders of magnitude, and Top-1 collapses from roughly $40\%$ to $24\%$. Computing the violation in cosine-margin space removes the loss spike, but it does not make hardest-negative probes beneficial.

Table~\ref{tab:cbrm_diagnostic} isolates the probe effect under a frozen backbone. BS-cRT, which uses no probes, reaches $41.07$ Top-1 and $25.06$ Few-shot accuracy over three seeds. Adding corrected hardest-negative probes gives similar Top-1, $40.86$, but reduces Few-shot accuracy to $20.70$, a drop of $4.36$ points. The random-negative row is a one-seed exploratory diagnostic rather than a full result; it does not show the same tail degradation, suggesting that the failure is tied to the systematic head anchoring induced by hardest-negative mining.

\begin{table}[t]
\centering
\caption{Frozen-backbone diagnostic on CIFAR-100-LT IF $100$. BS-cRT disables probes. The corrected CBRM variants compute violations in cosine-margin space. Values are percentages. Rows with $\pm$ report mean, and standard deviation over three seeds; the random-negative row is an exploratory one-seed diagnostic.}
\label{tab:cbrm_diagnostic}
\setlength{\tabcolsep}{4pt}
\small
\begin{tabular}{lrrrr}
\toprule
Stage-2 variant & Top-1 & Many & Medium & Few \\
\midrule
BS-cRT, no probes & $41.07\pm1.01$ & 57.16 & 40.99 & $25.06\pm0.95$ \\
CBRM, hardest negative & $40.86\pm0.99$ & 61.07 & 40.80 & $20.70\pm0.83$ \\
CBRM, random negative (1 seed) & 40.15 & 54.79 & 39.53 & 26.09 \\
\bottomrule
\end{tabular}
\end{table}

\begin{figure*}[t]
\centering
\includegraphics[width=0.95\textwidth]{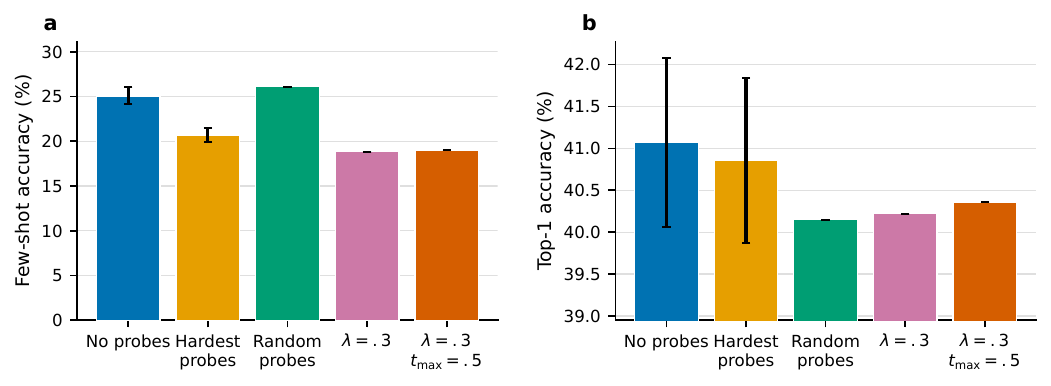}
\caption{Probe diagnostics on CIFAR-100-LT IF $100$. Panel (a) compares Few-shot accuracy across no-probe and probe-based stage-2 variants. Panel (b) shows the corresponding Top-1 accuracy. Error bars appear for variants with three seeds; one-seed diagnostic variants are shown without error bars. Lowering the CBRM weight or shortening the probe search interval does not recover the Few-shot accuracy of BS-cRT.}
\label{fig:probe_diagnostics}
\end{figure*}

Figure~\ref{fig:probe_diagnostics} expands this comparison with additional one-seed CBRM variants. Lowering the CBRM weight to $\lambda_{\mathrm{cbr}}=0.3$ or reducing $t_{\max}$ does not recover the tail accuracy of the no-probe baseline. The same pattern appears across these variants: the variant that omits synthetic probes is the most reliable stage-2 baseline, while hardest-negative probes tend to move accuracy back toward Many-shot classes at the expense of the tail.

These diagnostics identify the failure mode of the tested CBRM construction. Hardest-negative probes are intended to strengthen weak boundaries, but in long-tailed data, the hardest negative for a tail sample is frequently a head class. The generated probe is therefore anchored near a head-class region. Enforcing a positive tail margin on such probes can pull the tail classifier direction toward head-adjacent feature space, which is precisely the bias that classifier retraining is meant to reduce. In this setting, balanced classifier retraining without synthetic probes is the more reliable intervention.


\section{Discussion}

The results indicate that part of the residual tail error after Balanced Softmax remains in the classifier. After stage-1 training, the representation can still support a classifier whose directions reflect the long-tailed training stream. Retraining only the classifier under balanced class exposure recovers Few-shot accuracy in every evaluated benchmark cell, while aggregate Top-1 depends on how much Many-shot accuracy is traded for Medium- and Few-shot gains. This matters when interpreting aggregate accuracy: BS-cRT is not designed to improve all frequency groups uniformly. It changes the operating point of the classifier by reallocating decision capacity away from dominant classes and toward underrepresented classes during end-to-end training.

\subsection{Classifier-side correction after Balanced Softmax}

BS-cRT is intentionally limited in scope. It does not introduce a new architecture,
representation loss, auxiliary branch, or source of supervision. Its contribution is the controlled finding that a frozen representation trained with Balanced Softmax still benefits from a separate classifier-retraining stage. The use of matched stage-1  checkpoints is central to this claim: the comparison isolates the effect of the second-stage classifier update rather than mixing it with representation changes. In this setting, balanced $P \times K$ batches give each selected class comparable influence on the classifier update, while the empirical-prior Balanced Softmax loss keeps the same prior-aware objective used in stage 1.

This result has a methodological consequence for long-tailed recognition. Methods
that add prototypes, contrastive heads, generated samples, expert branches, or adaptive margins should be compared not only with end-to-end Balanced Softmax, but also with a matched classifier-retraining stage. Otherwise, part of the reported tail improvement may reflect the generic effect of re-estimating the final classifier under balanced exposure. BS-cRT is therefore best viewed as a low-complexity reference baseline rather than as a replacement for representation-learning or multi-stage long-tailed methods.

\subsection{Why hardest-negative boundary probes fail}

CBRM was motivated by a plausible hypothesis: tail classes have too few samples to
densely constrain their decision boundaries, so synthetic probes near current boundaries
might provide useful additional supervision. The experiments do not support this
extension under hardest-negative mining. The first failure mode is numerical. In a
normalized classifier, the margin should be measured in cosine-similarity units. Computing
the violation on scaled logits combines a cosine-range target margin with the classifier
scale, which destabilizes the boundary-risk term. Computing violations directly in
cosine-margin space removes the training collapse, but it does not make the probes
beneficial.

The remaining failure mode is geometric. In long-tailed data, the hardest negative for
a tail sample is often a head class. A probe constructed between a tail prototype and a
head-class prototype is therefore not a neutral sample near a balanced boundary. It is
anchored near a region already supported by many head-class examples. Enforcing a tail
margin at that location can pull the tail classifier toward head-adjacent feature space,
partially undoing the correction introduced by balanced classifier retraining. The
random-negative diagnostic is consistent with this interpretation: when negative classes
are sampled uniformly, the same tail degradation is reduced, although this one-seed
variant is not sufficient to define a new method.

These observations do not rule out boundary-aware learning for long-tailed recognition.
They indicate that boundary probes must account for class frequency, class support, and
classifier geometry. In particular, negative-class selection should avoid conflating
``hard'' with ``head-dominated,'' probe placement should account for asymmetric support
between head and tail classes, and the supervised margin should be expressed in the
same units as the classifier.

\subsection{Limitations and future work}

Several aspects of the study delimit the scope of the conclusions. First, the backbone family is fixed within each benchmark. We use ResNet-style architectures, but do not test wider convolutional networks, vision transformers, self-supervised pretraining, or foundation-model features.
The strength of BS-cRT may change when the stage-1 representation is learned under a
substantially different pretraining or architecture regime.

Second, all main benchmark cells use three seeds. The Few-shot gains are large relative to seed variation, but aggregate Top-1 changes should be interpreted more carefully, especially on ImageNet-LT, where BS-cRT trades aggregate accuracy for a stronger Few-shot operating point. Third, the stage-2 anchor loss is not ablated. All reported BS-cRT runs use empirical-prior Balanced Softmax under balanced class exposure. Replacing this anchor with cross-entropy, a uniform prior, or another prior choice may lead to different operating points.

Fourth, the paper focuses on accuracy and frequency-bucket behavior rather than
calibration. Since classifier retraining changes the distribution of confidence across head and tail classes, per-bucket reliability and post-hoc calibration should be studied separately. Fifth, iNaturalist 2018 is not included in the reported benchmark set, even though it is an important natural long-tailed dataset. Finally, the CBRM conclusions are specific to prototype-to-prototype interpolation with hardest-negative selection. Other forms of synthetic boundary supervision may still be effective if their probes are frequency-aware and geometrically aligned with tail-region expansion.

\section{Conclusion}

We studied classifier retraining after Balanced Softmax for long-tailed recognition.
The results show that a frozen-backbone stage-2 update can recover substantial
Few-shot accuracy from the same stage-1 representation. BS-cRT improves Few-shot
accuracy over the matched Balanced Softmax checkpoint across CIFAR-LT,
ImageNet-LT and Places-LT, without changing the inference architecture. The method 
therefore provides a useful low-complexity baseline for evaluating more elaborate
long-tailed recognition systems.

We also evaluated CBRM as a boundary-probe extension of this baseline. Correcting
the margin computation from scaled-logit space to cosine-margin space stabilizes
training, but hardest-negative probes still fail to improve over BS-cRT and can reduce tail accuracy. The analysis indicates that head-class anchoring is a central risk when synthetic boundary probes are generated by hard-negative mining in long-tailed data.
Future boundary-aware methods should therefore control how negative classes are
selected, where probes are placed, and how margins are defined relative to the classifier
geometry.



\end{document}